%% file: main.tex
\documentclass{article}

\usepackage{authblk}
\usepackage{comment,url}

\usepackage{subcaption}
\usepackage{type1cm}        
\usepackage{makeidx}         
\usepackage{graphicx}        
\usepackage{multicol}        
\usepackage[bottom]{footmisc}
\usepackage[nocomma]{optidef} 

\newtheorem{remark}{Remark}

\usepackage{newtxtext}       %
\usepackage[varvw]{newtxmath}       
\DeclareMathOperator{\sat}{sat}

\usepackage{float}
\usepackage{multirow}

\title{MILP initialization for solving parabolic PDEs with PINNs}
\author[1]{Sirui Li}
\author[2]{Federica Bragone}
\author[1,3]{Matthieu Barreau}
\author[1]{Kateryna Morozovska}
\date{January 2025}

\affil[1]{KTH Royal Institute of Technology, Malvinas v\"ag 10, 100-44, Stockholm, Sweden}
\affil[2]{KTH Royal Institute of Technology, Lindstedtsv\"gen 5, 100-44, Stockholm, Sweden}
\affil[3]{Digital Futures, Osquars Backe 5, 100 44 Stockholm, Sweden}

\begin{document}

%
%
\maketitle
\begin{abstract}
    Physics-Informed Neural Networks (PINNs) are a powerful deep learning method capable of providing solutions and parameter estimations of physical systems. Given the complexity of their neural network structure, the convergence speed is still limited compared to numerical methods, mainly when used in applications that model realistic systems. The network initialization follows a random distribution of the initial weights, as in the case of traditional neural networks, which could lead to severe model convergence bottlenecks. 
    To overcome this problem, we follow current studies that deal with optimal initial weights in traditional neural networks. In this paper, we use a convex optimization model to improve the initialization of the weights in PINNs and accelerate convergence. We investigate two optimization models as a first training step, defined as pre-training, one involving only the boundaries and one including physics. The optimization is focused on the first layer of the neural network part of the PINN model, while the other weights are randomly initialized. We test the methods using a practical application of the heat diffusion equation to model the temperature distribution of power transformers. The PINN model with boundary pre-training is the fastest converging method at the current stage.
\end{abstract}

\input{Introduction}

\input{Method}

\input{Results}

\bibliographystyle{plain}
\bibliography{Reference} 

\end{document}

%% file: Introduction.tex
\section{Introduction}
\label{sec:intro}
Physics-Informed Neural Networks or PINNs is an emerging deep learning method that combines a Feedforward Neural Network (FNN) and prior knowledge of the system expressed by Ordinary or Partial Differential Equations (ODEs and PDEs) \cite{raissi2019physics}. With its booming popularity in research, PINN has been applied to a variety of problems across many scientific domains such as physics of fluids \cite{almajid2022prediction, wessels2020neural}, dynamical systems \cite{linka2022bayesian, antonelo2024physics}, energy systems \cite{misyris2020physics}, heat diffusion \cite{Fede2, Fede3} and many more.  

Many problems mentioned above are tested using simplified system representation to showcase the potential benefits of PINN in the future. It is expected that PINNs will be able to compete with traditional numerical methods for applications where the need for fast convergence surpasses the need for aiming for exact solutions. Therefore, potentially, PINNs can cover the gap in applications, which can allow for a small degree of uncertainty but require methods with faster speed of convergence and smaller availability of computing power. However, in the current state-of-art on PINN applications, we still observe convergence speed, which is slower than potentially desired, and it increases exponentially with scaling models to represent realistic scenarios. 

Similarly to other artificial neural networks, PINNs initialization starts with randomly distributed initial weights, which are then fitted to represent the problem better along with the training process. Therefore, depending on the initialization parameters, the convergence rate can be significantly different, and as described in \cite{goodfellow2016deep}, poor initial weights can create bottlenecks in model convergence and potentially result in the model not being able to converge because of the weights instability.

Different activation functions use various techniques to obtain initial weights. Currently, the most common initial weighting techniques for the activation functions $tanh$ and $sigmoid$ are the 'Glorot' or 'Xavier' initialization, proposed by Xavier Glorot and Yoshua Bengio in 2010 \cite{glorot2010understanding}. This common random initialization technique initializes the weights according to the Glorot Normal distribution with the variance given by \eqref{glorot}.
\begin{equation}
    var(W)=\frac{2}{n_{in}+n_{out}},
    \label{glorot}
\end{equation}
where $n_{in}$ and $n_{out}$ are the number of input and output units, respectively. Random uncertainty can still lead to a poor initial training starting point and excessively long training times. 

Since the structure of the PINN contains penalty information given by prior knowledge, its convergence difficulty is higher than that of the ordinary FNN. As the amount of information the neural network model needs to process increases, this convergence problem will become more apparent. For example, in applications related to heat diffusion, the shift from 1D spatial representation to 2D and 3D systems causes the need for more data as initial and boundary conditions as well as collocation points for training, and this data increase grows exponentially with additional domains and complex geometries. 

As the development of PINNs moves forward, several works show possible flaws in the training of FNNs used in the PINNs architecture and propose new algorithms to improve the performance and accuracy of the training. In \cite{wang2021understanding}, the authors show unbalanced gradients during model training using back-propagation,  which is related to stiff gradient flow dynamics. They mitigate this problem by introducing a learning rate annealing algorithm to adapt the assigned weights of the PINNs' loss function terms. In \cite{wang2022and}, the authors derive the Neural Tangent Kernel (NTK) of PINNs, which describes the training evolution of a network with infinite-width \cite{jacot2018neural}, and derive its convergence. Showing that PINNs suffer from spectral bias, which limits their learning of higher frequencies \cite{rahaman2019spectral}, and convergence rate in the loss function terms, they propose a novel algorithm to overcome these problems.
In \cite{barreau2021physics}, to deal with several constraints, the authors suggest improving the training procedure by penalizing unfeasible solutions of the interested system by applying an extended Lagrangian cost function. Their training procedure is split into two parts: the pre-training using ADAM optimizer and updating the weights assigned to the loss functions, and the training step where they fixed the weights and used a second-order optimizer like BFGS.
 Other techniques to improve and fasten the convergence of PINNs include domain composition, like Conservative Physics-Informed Neural Networks (CPINNs) \cite{jagtap2020conservative} and Extended Physics-Informed Neural Networks (XPINNs) \cite{jagtap2020extended}. 
However, these solutions still require larger computational resources and can be improved by additional approaches. Therefore, in this study, we aim to use convex optimization to find more suitable initial training weights, which could allow for faster and more resource-effective training of the PINN model. The optimization model follows the approach described in \cite{Optimization-tanh_main}. While the authors apply their method to generic FNNs, we extend the work to PINNs. Other work, like Extreme Learning Machine \cite{huang2006extreme}, focuses on random initialization and consequent optimization of the features and outputs. The Extreme Learning Machine was then extended to PINNs in \cite{dwivedi2020physics}, where the authors introduced the Physics-Informed Extreme Learning Machine (PIELM) as a novel and fast learning scheme for PDEs.
As for \cite{barreau2021physics}, we introduce a pre-training step where, in this case, the optimal initial weights are found. We tested both using only boundary conditions and then added the physical part. Afterward, we train the PINN using the optimal weights.

In this paper, we propose using a mathematical optimization model to facilitate the training of PINN and accelerate its convergence. The optimization model aims to express the mathematical relaxation model of the entire PINN and use an optimization algorithm to find the optimal weights and biases suitable for the PINN. Due to the relaxation, the weights and biases given by the optimization model are the bounds of PINN's optimal weights and biases rather than the final optimization results. They are used to provide direction for PINN training and reduce the uncertainty caused by random initial weights and biases. This work applies parabolic PDE and uses the heat transfer problem inside a power transformer as a practical example. Considering the structural complexity of deep neural networks, we center the optimization on the first layer and keep the other layers the same as the original PINN, i.e., using randomly selected initial weights.

%% file: Method.tex
\section{Method}
\label{sec: method}

This section outlines the methodology for this study including the test case, which was used for testing the performance of PINN with and without optimal weight initialization. The test case is based on an application of heat diffusion equation to a problem of heat distribution in power transformer \cite{Fede2, Fede3}. The input data consists of measurements of top and bottom temperatures of the power transformer, which is filled with mineral oil. The transient heat diffusion problem is further expressed in a form of PDE and later used as a basis for constructing PINN and an optimization model for finding optimal initial weights. 

\subsection{PINNs}
\label{subsec:pinns}

PINNs are generally constructed using feedforward neural network in combination with known function in a form of ODE or PDE which is integrated in the neural network loss function. 

In this study, we will consider the following neural network with only one hidden layer:
\begin{equation}
    \hat{u}(X) = W_2 \sigma(W_1 X + b_1) + b_2,
    \label{eq: FNN output}
\end{equation}
where $\sigma$ is the activation function for the hidden layer; $W_1$ and $W_2$ are the weights matrices; $b_1$ and $b_2$ are the biases; $X$ is the input vector; $\hat{u}$ is the output. To simplify the writing, we assume that $\Theta = \left\{ (W_i, b_i) \right\}_{i=1,2}$ is the tensor of parameters.


The overall loss function in PINN is defined as the weighted sum of the mean-squared error for the boundary conditions, MSE$_u$, and the mean-squared error for the residual operator $f$, MSE$_f$, given by:
\begin{equation} 
    \mathfrak{L}_{\lambda_u, \lambda_f, f}(\Theta) = \lambda_u \text{MSE}(\hat{u}(X_u), u) + \lambda_f \text{MSE}(f[\hat{u}](X_f), 0), \label{eqMSE}
\end{equation}
where 
\[
    \text{MSE}(v, w) = \frac{1}{N} \sum_{i=1}^{N}{|v_i - w_i|^2},
\]
$N$ being the number of rows in both $v$ and $w$. In that work, we consider $\{X_u^{i}\}_{i=1}^{N_u}$ as the inputs generating the measured outputs $\{u^{i}\}_{i=1}^{{N}_u}$; and $\{X_f^j\}_{j=1}^{{N}_f}$ is the collocation points\footnote{This points are generated uniformly and randomly inside the domain before the training.}; $N_u$ is the number of boundary points; $N_f$ is the number of collocation points, and $\{ \lambda_u, \lambda_f\}$ are the weights assigned to each loss function.


\subsection{Heat diffusion equation}

The general form of the heat diffusion equation in 1D is given by:
\begin{equation}
      \rho c_pu_t = ku_{xx} + q.
    \label{eq:heat_diff}
\end{equation}
In the equation, $\rho$ is the density, $c_p$ is the specific heat capacity, and $k$ is the thermal conductivity. The heat source $q$ used in this study is picked to model the temperature distribution of a power transformer, and it is defined as follows:
 \begin{align}
     & q = q(x,t) = (P_0 + P_K(x,t) - h(u(x,t) - T_a(t))),\\
     & P_K(x,t) = \nu K(t)^2(0.5\sin(3 \pi x) + 0.5),
 \end{align}
 where $P_0$ is the no-load loss, $P_K(x,t)$ is the load loss depending on the load factor $K(t)$ and the rated load loss $\nu$, $h$ is the convective heat transfer coefficient, and $T_a(t)$ is the ambient temperature as a function of time.
The boundary conditions are also specific to the problem and are defined as:
\begin{align}
    & u(x_0,t) = T_a, \\
    & u(x_{end},t) = T_{o},
\end{align}
where $T_o$ is the top-oil temperature. A list of the used values in the equation with corresponding units is given in Table \ref{tab:phys_param}. Our data consists of measurements from an in-operation transformer for $T_a$, $T_o$, and $K$.

\begin{table}[h]
\renewcommand{\arraystretch}{1.3}
\centering
\caption{Physical parameters and corresponding values of the considered heat diffusion equation.}
\begin{tabular}{|c|c|c|}
\hline
\textbf{Parameter} & \textbf{Value} & \textbf{Unit}\\
\hline
Thermal conductivity, $k$   & 50 & $[W/mK]$    \\ \hline
Density, $\rho$  & 900 & $[kg/m^3]$  \\ \hline
Specific heat capacity, $c_p$  & 2000 & $[J/kgK]$ \\ \hline
Convective heat transfer coefficient, $h$    & 1000 & $[W/m^2K]$ \\ \hline
No-load loss, $P_0$  & 15000 & $[W]$    \\ \hline
Rated load loss, $\nu$  & 83000 & $[W]$    \\ \hline
\end{tabular}
\label{tab:phys_param}
\end{table}
 
\begin{figure}
    \centering
    \includegraphics[width=0.9\linewidth]{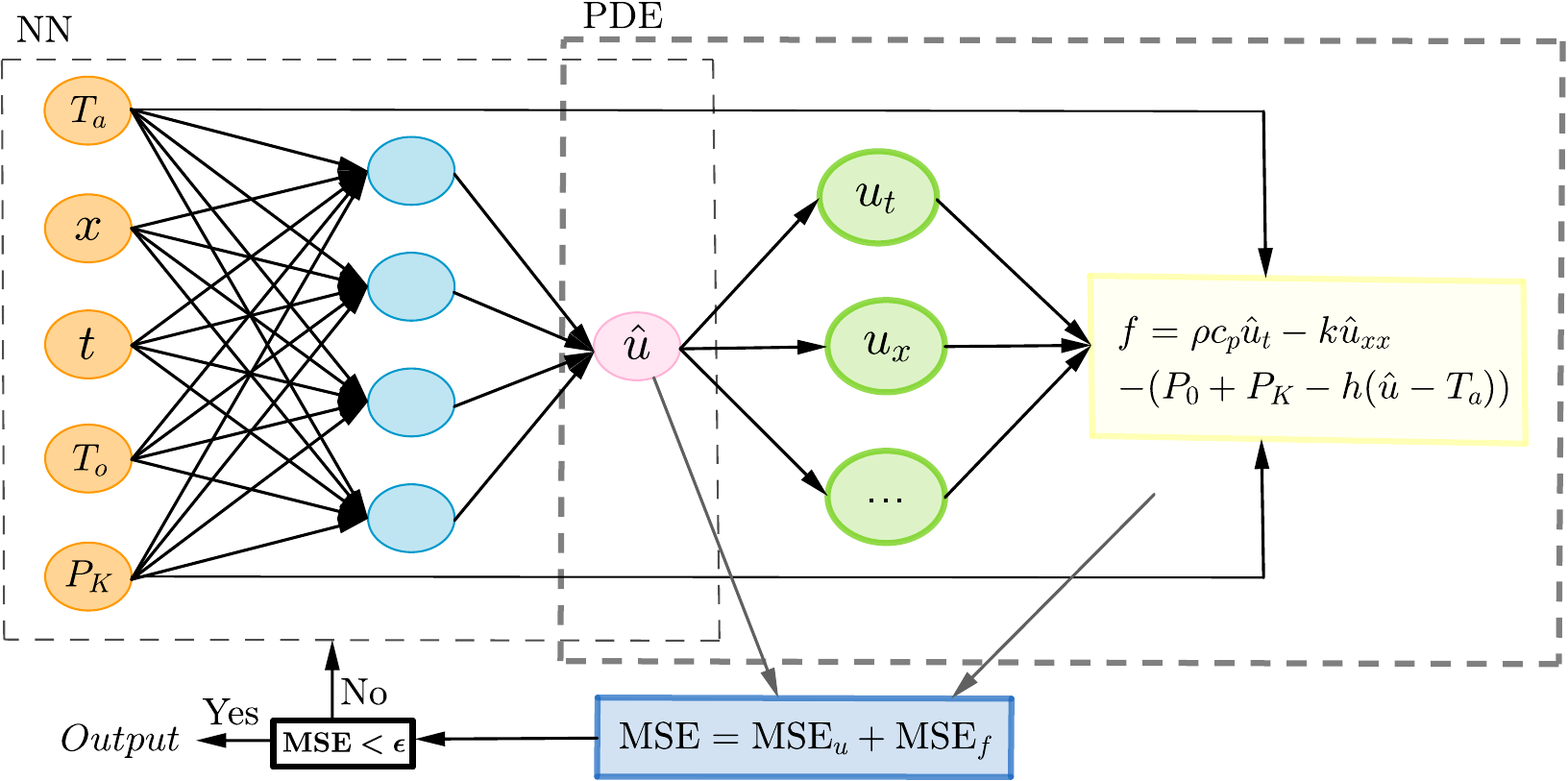}
    \caption{Structure of PINN for the heat diffusion equation.}
    \label{fig:PINN structure}
\end{figure}

We define the residual function $f$ for Eq. \eqref{eq:heat_diff} as follows:
\begin{equation}
     f[u](X) = 
     c_pu_t - ku_{xx} -(P_0 + P_K - h(u - T_a)).
     \label{eq: f(Xf)}
\end{equation}
In a PINN framework, we use the feedforward neural network \eqref{eq: FNN output} with 
\[
    X = \left[ \begin{matrix} T_a & x & t & T_o & P_K \end{matrix} \right]^{\top}
\]
as the input vector and the output will be the estimated temperature $\hat{u}$ at location $x$, time $t$ with the external inputs $T_a, T_o$ and $P_K$. A schematic representation of PINN used to model heat diffusion problem in this study is outlined in Figure \ref{fig:PINN structure}. 

\subsection{Optimization model formulation}
In order to reduce the time needed for training a PINN model and increase its accuracy we use a mixed integer linear optimization to find better initial weights for training the model. The optimization model focuses on optimizing the neural network loss function similarly to the approach described in \cite{Optimization-tanh_main}.
 Traditionally, we optimize the parameters $\Theta$ by minimizing the loss function:

\begin{align}
\label{model: FNN}
    \min_{\Theta} & \displaystyle \frac{1}{N} \sum_{i=1}^N |\hat{u}(X_i) - u_i|^2,
\end{align}
where $\{X_i\}_{i=1 \dots N}$, $\{u_i\}_{i=1 \dots N}$ is the training dataset. In the heat diffusion case, $X_i$ is the time, position, and external inputs leading to the temperature measurement $u_i$.

The problem presented is non-linear and non-convex, which makes it difficult to solve the optimization model. 
By expanding the network size with more hidden layers, the complexity of the loss function in Eq. \eqref{model: FNN} increases. 

In this work, we want to use Mixed-Linear Programming (MLP) to find an approximate solution before training the network. Therefore, we propose to use a piecewise linear approximation of the activation function. In our case, we choose $\sigma(x) = \tanh(x)$ and the following approximation:
\begin{equation}
    \sigma(x) = \tanh(x) = \frac{2}{1+e^{-2x}}-1 \approx \sat (x) = \begin{cases}
        -1, & \text{if } x < -1\\
        x, & \text{if } -1 \leq x \leq 1 \\
        1, & \text{if }  x > 1
    \end{cases}
    \label{eqLinearize}
\end{equation}

\begin{figure}
    \centering
    \includegraphics[width=0.95\linewidth]{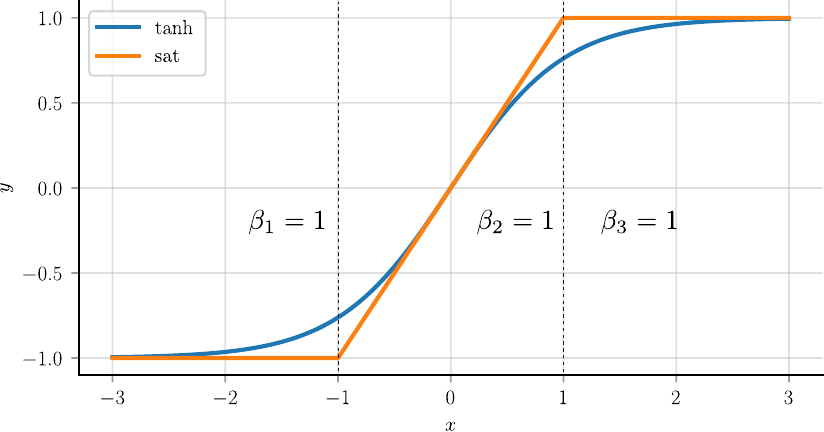}
    \caption{Comparison between the hyperbolic tangent and the saturation function. The $\beta_i$ variable is used for the translation to MLP language.}
    \label{fig:saturation}
\end{figure}

The quality of the approximation is depicted in Figure~\ref{fig:saturation}.

Since \eqref{eq: FNN output} is a single layer neural network with bounded input $X$, that means $x = W_1 X + b_1$ is also bounded. Then, we can assume that there exists $M$, potentially very large, such that $x \in [-M, M]$. There are three possibilities:
\begin{enumerate}
    \item If $x \in [-M, -1]$, we define this zone as $\beta_1 = 1, \beta_2 = \beta_3 = 0$. Then there exists $\gamma_1 \in [0, 1]$ such that $x = \gamma_1(x) M - \gamma_2(x)$ with $\gamma_2(x) = 1 - \gamma_1(x)$. In that case, we get
    \[ 
        \sat(x) = - 1 = - \gamma_1(x) - \gamma_2(x).
    \]
    
    \item If $x \in [-1, 1]$, we define this zone as $\beta_2 = 1, \beta_1 = \beta_3 = 0$. Then, there exists $\gamma_2 \in [0, 1]$ such that $x = -\gamma_2(x) + \gamma_3(x)$ with $\gamma_3(x) = 1 - \gamma_2(x)$. In that case, we get 
    \[
        \sat(x) = x = - \gamma_2(x) + \gamma_3(x).
    \]
    
    \item Finally, if $x \in [1, M]$, we define this zone as $\beta_3 = 1, \beta_1 = \beta_2 = 0$. Then, there exists $\gamma_3 \in [0, 1]$ such that $x = \gamma_3(x) + M \gamma_4(x)$ with $\gamma_4(x) = 1 - \gamma_3(x)$. In that case, we get 
    \[
        \sat(x) = 1 =  \gamma_3(x) + \gamma_4(x).
    \]
\end{enumerate}
Combining all these considerations lead to the following MLP problem:
\begin{align}
        & \sat(x) = - \gamma_1(x) - \gamma_2(x) + \gamma_3(x) + \gamma_4(x), \\
        & x = -M \gamma_1(x) - \gamma_2(x) + \gamma_3(x) + M \gamma_4(x),  \\
        & 0 \leq \gamma_1(x) \leq \beta_1(x), \\
        & 0 \leq \gamma_2(x) \leq \beta_1(x) + \beta_2(x), \\
        & 0 \leq \gamma_3(x) \leq \beta_2(x) + \beta_3(x), \\
        & 0 \leq \gamma_4(x) \leq \beta_3(x), \\
        & \gamma_1(x) + \gamma_2(x) + \gamma_3(x) + \gamma_4(x) = 1,  \\
        & \beta_1(x) + \beta_2(x) + \beta_3(x) = 1, \quad \beta_1(x), \beta_2(x), \beta_3(x) \in \{0,1\}.
\end{align}
As written above, for each $x$, we must define new variables $\gamma_i$ and $\beta_i$ which means that for each $x$ investigated, we need to create $7$ variables to compute $\sat(x)$.


Moreover, we introduce a change in the objective function to address the non-linearity issue. Instead of using the MSE loss function, as introduced in Sec. \ref{subsec:pinns}, we aim to minimize the Mean Absolute Error (MAE). The reason is that MSE works great when using gradient descent schemes since its gradient is proportional to the error made \cite[Chap. 8]{Goodfellow-et-al-2016}. However, when using MLP, MAE is better suited because it can be linearized. This adjustment might lead to slightly different final optimal values for weights and biases, but MAE and MSE will tend to converge to the same optimal solution of $0$ as the number of neurons increases.


Taking the PINN model with one hidden layer as an example, the new objective function thus is
\begin{equation}
    \mathcal{L}_{\lambda_u, \lambda_f, f}(\Theta) = \lambda_u MAE(\hat{u}, u) + \lambda_f MAE(f[\hat{u}], 0),
\end{equation}
where 
\[
    MAE(v, w) = \frac{1}{N} \sum_{i=1}^{N} |v_i - w_i|
\]
where $N$ is the number of lines of both $v$ and $w$. We consider here the dataset to consist of $\{X_i, u_i\}_{i=1, \dots, N_u}$ for real measurements and $\{X_j\}_{j=1,\dots,N_f}$ for physics collocation points (fictitious points at which the physical model \eqref{eq:heat_diff} should hold).

In our case, we consider using a fixed value matrix $W_2$. This means that the pre-training will try to find the best features to use. Compared to other strategies such as Extreme Learning Machine \cite{huang2006extreme} where they use random features and optimize to find the best combination of these random features, we expect this method to be better suited to PINNs. 

We will consider two optimization procedures as pre-training. The first one, named "boundary pre-training", consider to pre-train $W_1, b_1$ and $b_2$ such that $\hat{u}$ is fitting some boundary points already. This writes as
\begin{align}
    \min_{W_1,b_1,b_2} \quad & \mathcal{L}_{1,0,0}(\Theta) \label{rm1}\\ 
    s.t. \quad
    & \text{\eqref{eq: FNN output} holds.}
\end{align}

The second optimization problem we will consider incorporating the physics-loss to investigate how a fitting with additional in-domain knowledge would improve or not the previously defined model. As of with PINN, we can incorporate the physical knowledge as
\[
    f[\hat{u}] = 0
\]
where $f$ is defined in Equation~\eqref{eq: f(Xf)}. The issue is that $f$ is an operator which includes derivatives of $\hat{u}$ with respect to time and space. This operation will necessarily bring non-linearity in $\mathcal{L}_{\lambda_u, \lambda_f, f}$ by multiplying the weight matrices $W_1$ and $W_2$ together. The most elegant solution is instead to approximate the derivatives by a finite difference operator 
\[
    \hat{u}_t(t, x) \simeq \frac{1}{\varepsilon}\left(\hat{u}(t + \varepsilon, x) - \hat{u}(t, x)  \right).
\]
By doing the same operation on the space derivative as well, we get the operator $\tilde{f}_{\varepsilon}[u] \sim_{\varepsilon \to 0} f[\hat{u}]$. This operator $\tilde{f}$ is however linear in $\Theta$ as long as the original $f$ is linear in $u$. This is our case in that problem, leading to a linear formulation of $\mathcal{L}_{\lambda_u,\lambda_f,\tilde{f}}$.

The optimization problem which contains the physical information, denoted "full pre-training", is then derived as
\begin{align}
\label{model: CM}
    \min_{W_1,b_1,b_2}  \quad &   \mathcal{L}_{\lambda_u, \lambda_f, \tilde{f}}(\Theta) 
    \\ 
    s.t.  \quad
\end{align}
where $X_f = [T_a, x_f, t_f, T_o, P_K]$ are the training points for the residual $f$; $q_u$ and $q_f$ can be defined as activations, and are linear combinations of the input $X_f$.

\begin{remark}
    Note that the approximation we made of $f$ is quite crude since we use $\tanh = \sat$ to compute the derivative. It would have been better to approximate the derivative of the hyperbolic tangent $\tanh' = 1 - \tanh^2 \simeq 1 - \sat^2$, which is more accurate, but the resulting operator $f$ would then be nonlinear. In order to isolate the effect of the physics loss on the final solution with optimal weights, we can compare these two models. We discuss the impact of this approximation in the following section.
\end{remark}

These two models are investigated as a first training step, denoted pre-training. The second step is the optimization of the overall problem using the classical PINN procedure, i.e., approximating a solution to 
\begin{equation}
    \label{eq:PINN_training}
    \min_{\Theta} \mathcal{L}_{\lambda_u, \lambda_f, f}(\Theta)
\end{equation}
using first a gradient-descent algorithm (ADAM) and then a second-order solver such as BFGS. The complete training methodology is described in the following section.

%% file: Results.tex
\section{Results and Discussion}
\label{sec: result}


The neural network structure for the PINN model consists of 5 neurons in the input layer, one hidden layer, and 1 neuron in the output layer, for a total of 3 layers. We tested different numbers of neurons for the hidden layer, showing the results obtained with 32 and 60 nodes. To ensure stability in the convergence of the PINN, we standardized the input data and normalized the output. We use the "Xavier" weights initialization and the $\tanh$ as the activation function. We used both ADAM optimizer with a learning rate of 1e-2 and L-BFGS-B. For more details about the model specifications, refer to previous studies \cite{Fed, Fede2, Fede3}. We first investigate the outputs of the pre-training models, then compare the convergence speed, and end with an evaluation of the performance.


\subsection{Pre-training evaluation}

\begin{figure}[H]
\centering
     \begin{subfigure}[b]{0.9\textwidth}
         \centering
         \includegraphics[width=\textwidth]{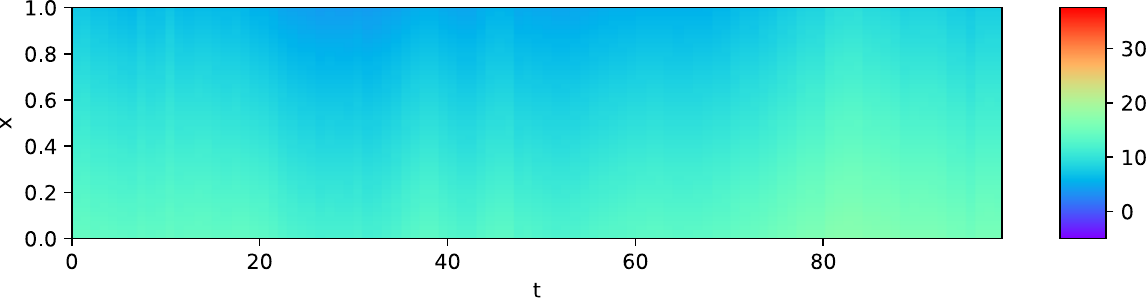}
         \caption{Vanilla PINN.}
         \label{fig:orig_pinn_32_init}
     \end{subfigure}
          \hfill
     \begin{subfigure}[b]{0.9\textwidth}
    \centering
    \includegraphics[width=\textwidth]{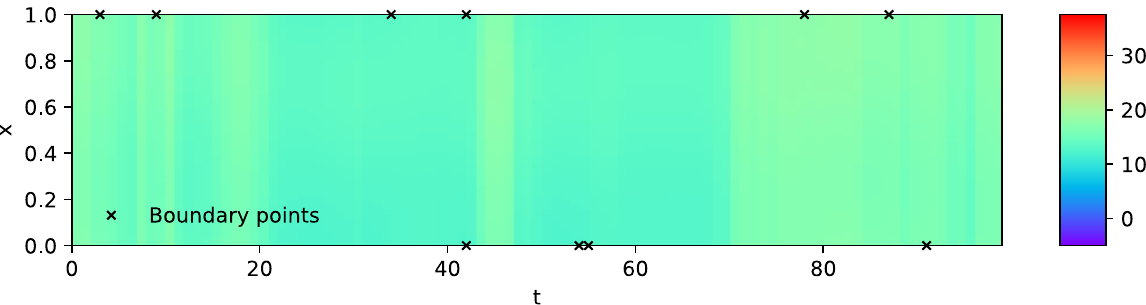}
         \caption{PINN with boundary pre-training.}
         \label{fig:model1.1_32_init}
    \end{subfigure}
        \hfill
     \begin{subfigure}[b]{0.9\textwidth}
    \centering
    \includegraphics[width=\textwidth]{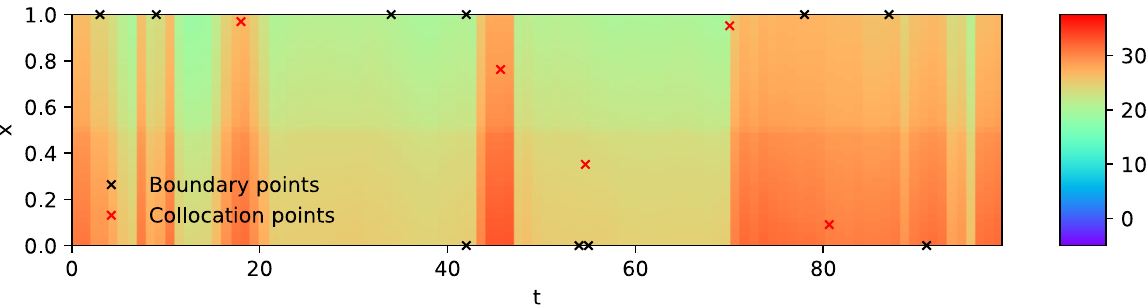}
         \caption{PINN with full pre-training.}
         \label{fig:model1_32_init}
        \end{subfigure}

    \caption{Temperature solution given by the initial weights of the vanilla PINN and PINN using boundary and full pre-training with 32 neurons.}
\label{fig:init_weights_all_32}
\end{figure}

\begin{figure}[H]
\centering
     \begin{subfigure}[b]{0.9\textwidth}
    \centering
    \includegraphics[width=\textwidth]{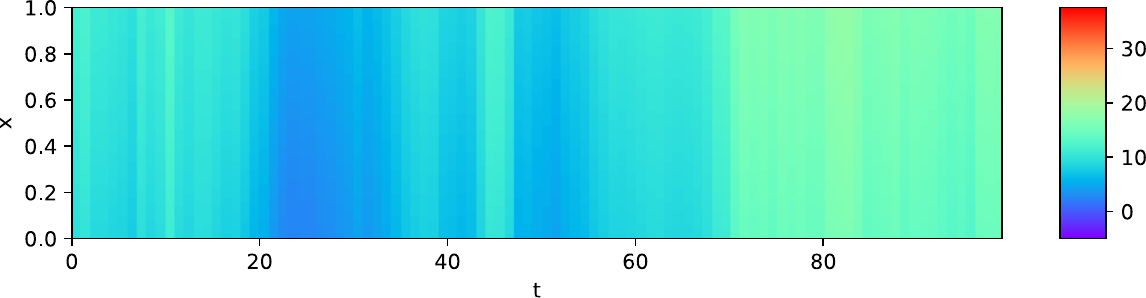}
         \caption{Vanilla PINN.}
         \label{fig:orig_pinn_60_init}
    \end{subfigure}
        \hfill
     \begin{subfigure}[b]{0.9\textwidth}
    \centering
    \includegraphics[width=\textwidth]{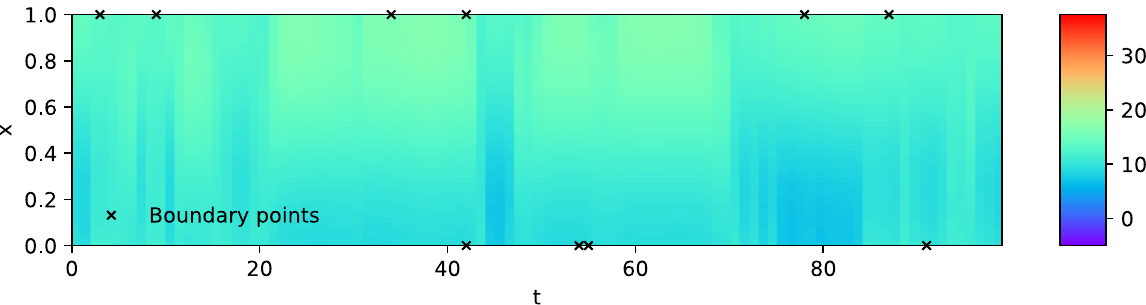}
         \caption{PINN with boundary pre-training.}
         \label{fig:model1.1_60_init}
    \end{subfigure}
             \hfill
     \begin{subfigure}[b]{0.9\textwidth}
    \centering
    \includegraphics[width=\textwidth]{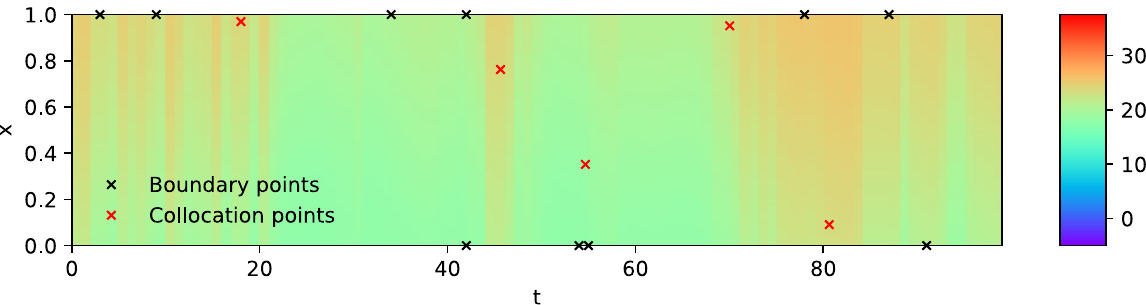}
         \caption{PINN with full pre-training.}
         \label{fig:model1_60_init}
    \end{subfigure}
    
\caption{Temperature solution given by the initial weights of the vanilla PINN and PINN using boundary and full pre-training with 60 neurons.}
\label{fig:init_weights_all_60}
\end{figure}

\begin{figure}
    \centering
    \includegraphics[width=0.9\linewidth]{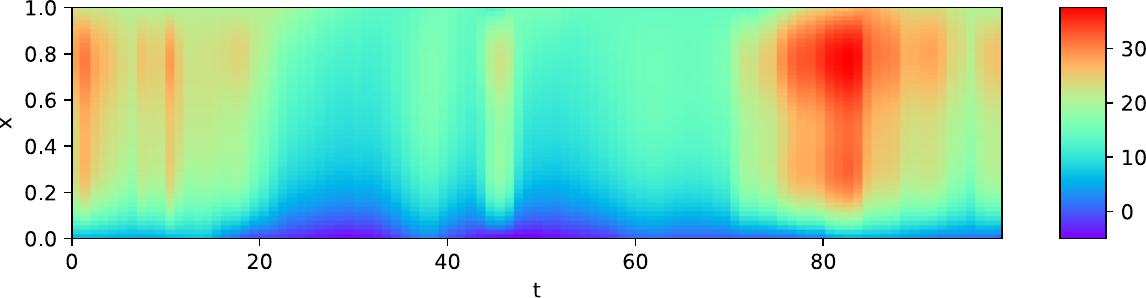}
    \caption{Comsol reference solution.}
    \label{fig:comsol_init}
\end{figure}

We start by analyzing the optimization models' initial weights and comparing them to the vanilla PINN model. We randomly select ten samples from the boundary training set for both optimization models to determine the optimal initial weights and biases. These ten points are the same for both models. For PINN using full pre-training, an additional five points inside the domain, i.e., collocation points, are randomly picked. We test our models to predict the temperature for a $50$ spatial points grid and $100$ time points. By plotting the temperature solution using only the initial weights, we can notice some first improvements in the optimized models, as shown in Figs. \ref{fig:init_weights_all_32} and \ref{fig:init_weights_all_60} for 32 and 60 neurons in the hidden layer for all the PINN models, respectively. Fig. \ref{fig:comsol_init} gives the reference solution obtained with Comsol. The models at this stage only use the initial weights, so the plots do not match the reference solution. However, we want to emphasize how the PINN models initialize their weights and reach different temperature ranges. 

Figs. \ref{fig:orig_pinn_32_init} and \ref{fig:orig_pinn_60_init} represent the temperature distribution given by the initial weights of the vanilla PINN using 32 and 60 neurons, respectively. We can notice that both plots start to capture some temperature changes, particularly for the model using 60 neurons, where we can see more distinguishable temperature regions closer to the reference solution. However, the temperature range is still lower than the final solution. The vanilla PINN model does not capture this wider temperature range because of the randomness of the weights initialization, which can produce uncertain initial results leading to consequent suboptimal predictions. Figs. \ref{fig:model1.1_32_init} and \ref{fig:model1.1_60_init} display the results of the initial weight using the optimized PINN model considering only the boundary points, i.e., PINN with boundary pre-training. This model has already slightly improved in terms of temperature range compared to the original PINN. The PINN model with full pre-training, where we are also considering five collocation points, improves the results further, spanning a range of temperatures closer to the reference solution, as it can be seen in Figs. \ref{fig:model1_32_init} and \ref{fig:model1_60_init} for the two considered networks architectures. 

Looking at the temperature distributions obtained by the two optimal models, we notice more discontinuities than in the vanilla PINN predictions. This is because we are using only 10 data points in the boundaries for both optimization models and five additional collocation points for the full pre-training model for training, which is insufficient to represent the entire domain. Moreover, the jumps in the model approximation of $f$ (see Remark~1) are naturally creating discontinuities.

\subsection{Convergence speed}

\subsubsection{Evaluation of the pre-training algorithms} 

We ran the optimization models 5 times and took the average time for both the boundary and full pre-training using 32 and 60 nodes. The boundary pre-training model takes an average of 5.196 seconds with 32 nodes and 5.938 seconds with 60 nodes. For the full pre-training, the average time is 32.996 seconds and 128.090 seconds for 32 and 60 nodes, respectively. For all the models, the variance is quite high, but it can be noticed that boundary pre-training takes relatively less time than full pre-training. 

\subsubsection{Evaluation of the training algorithm}

\begin{figure}
    \centering
    \includegraphics[width=0.9\linewidth]{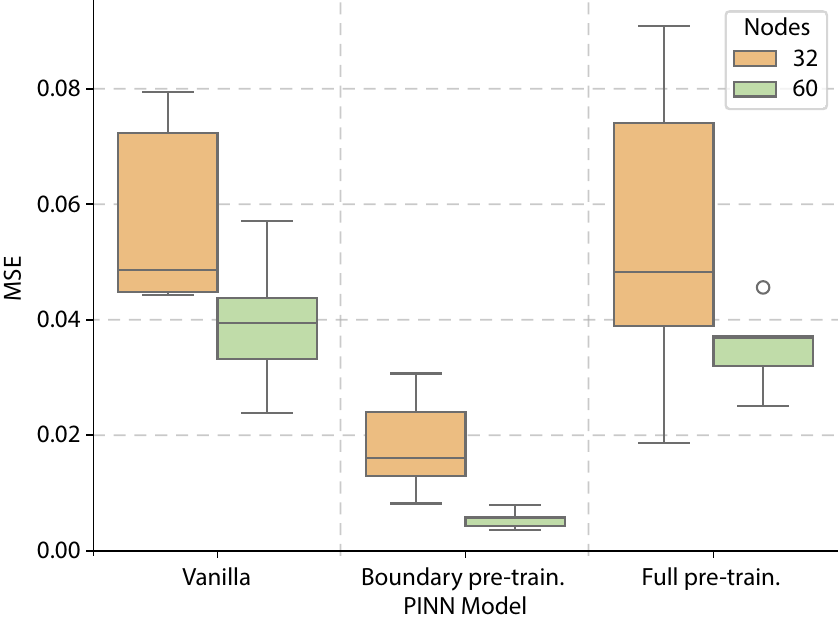}
    \caption{Box plot for the MSE values reached by vanilla PINN, PINN with boundary pre-training, and PINN with full pre-training for 32 and 60 nodes in 5 test cases.}
    \label{fig:boxplot_mse}
\end{figure}

We now focus on understanding the time it takes to run the proposed optimized models compared to the vanilla PINN. All the models are run for 5000 epochs using ADAM optimizer, and we report the loss value for each of them and the time and repeat five times. Fig. \ref{fig:boxplot_mse} shows the box plot for the total MSE the 3 models converged to after 5000 epochs in the 5 test runs for 32 nodes, in orange, and 60 nodes, in green.
Using 32 nodes, the average loss value for vanilla PINN is $0.0579\pm0.0150$ with an average time of 5:20 minutes. The loss function averages $0.0184\pm0.0081$ and $0.05416\pm0.0256$ for the PINN models with boundary pre-training and full pre-training in 5:21 and 5:23 minutes, respectively. As expected, the training time for all 3 models is similar since we are running for the same number of epochs. However, the PINN model with boundary pre-training reaches lower MSE values and has a lower variance between the 5 test runs than the other two models, as shown in Fig. \ref{fig:boxplot_mse}.
Using 60 nodes, the vanilla PINN reaches, on average, $0.0395\pm0.0111$ with an average time of 5:22 minutes. For PINN with boundary pre-training, the average loss value it attains is $0.0059\pm0.0015$ with an average time of 5:28, while for PINN with full pre-training, the average loss value attained after 5000 epochs is $0.0353\pm0.0067$ in 5:24 minutes. As for the case of 32 nodes, the running times are very similar. However, we can see a significant improvement using the PINN model with boundary pre-training, as it reaches an average loss of one magnitude lower than the other two models. This indicates that the model using 60 nodes would reach convergence much faster for a complete training run. In general, for the three models using 60 nodes, we notice less variance between the 5 test runs compared to the cases using 32 nodes, as shown in Fig. \ref{fig:boxplot_mse}, particularly for the PINN model with boundary pre-training.

\subsection{Evaluation of the training}

For the last part of our testing, we run the models for 10000 epochs with ADAM and then with L-BFGS-B until convergence. Table \ref{tab:accuracy_all} shows the six different runs for the 3 models using 32 and 60 nodes. We report the duration and the value of the total MSE they reached at convergence. Overall, all the models get a significantly low value for the total MSE. PINN with boundary pre-training performs the best as the training time is considerably lower, around 20 minutes, compared to approximately 30 minutes for both the vanilla PINN and the PINN with full pre-training.

\begin{table}[ht]
\caption{Duration and total MSE reached upon convergence using 10000 epochs with ADAM and L-BFGS-B until convergence afterward for vanilla PINN, PINN with boundary pre-training, and PINN with full pre-training with 32 and 60 nodes.}
\centering
\label{tab:accuracy_all}
\renewcommand{\arraystretch}{1.5}
\begin{tabular}{|p{2.4cm}|p{1.2cm}|p{0.8cm}|p{1.2cm}|p{0.8cm}|p{1.2cm}|p{0.8cm}|}
 \hline 
 & \multicolumn{2}{|p{2.5cm}|}{\centering \textbf{Vanilla \\ PINN}} & \multicolumn{2}{|p{2.5cm}|}{\centering \textbf{ Boundary pre-training PINN }} & \multicolumn{2}{|p{2.5cm}|}{\centering \textbf{ Full \\ pre-training PINN }} \\ 
 \hline
 \textbf{Nodes} & 32 & 60 & 32 & 60 & 32 & 60\\
 \hline
\textbf{Duration} (min) &  35:38 & 31:52 &  20:02 &  21:58 & 35:56 & 29:27\\
\hline
\textbf{Total MSE} (e-5) & 2.22306 & 0.52098 & 0.83575 & 0.75401 & 5.37374 & 0.73301 \\
 \hline
\end{tabular}
\end{table}

\begin{figure}[H]
\centering
     \begin{subfigure}[b]{0.9\textwidth}
    \centering
    \includegraphics[width=\textwidth]{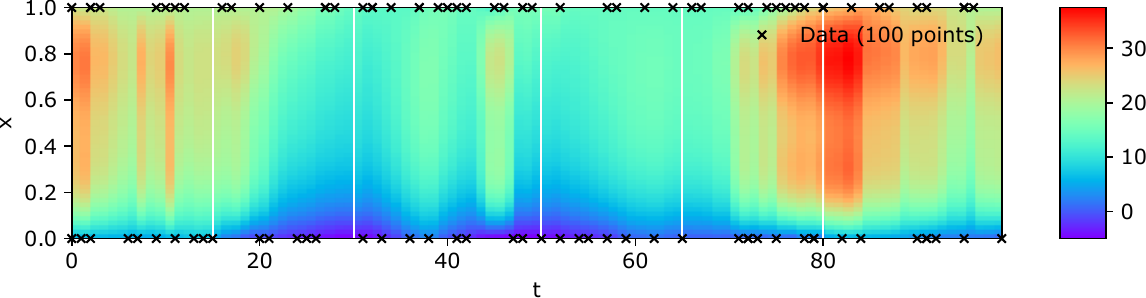}
         \caption{Vanilla PINN.}
         \label{fig:vanilla_pinn_32}
    \end{subfigure}
             \hfill
     \begin{subfigure}[b]{0.9\textwidth}
    \centering
    \includegraphics[width=\textwidth]{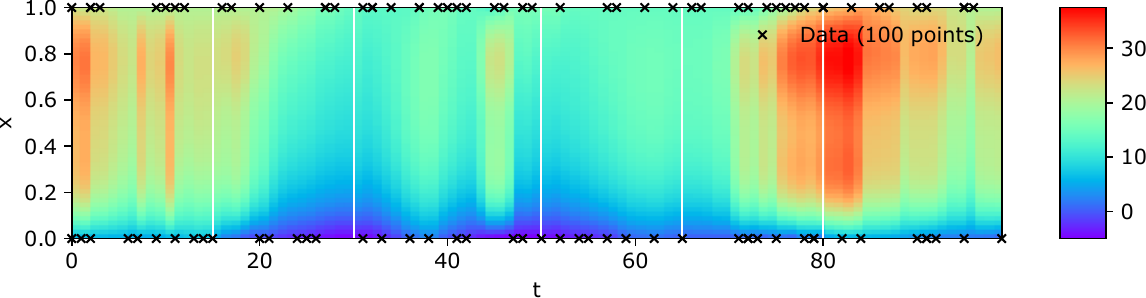}
         \caption{PINN with boundary pre-training.}
         \label{fig:pinn_boundpretrain_32}
    \end{subfigure}
    \hfill
     \begin{subfigure}[b]{0.9\textwidth}
    \centering
    \includegraphics[width=\textwidth]{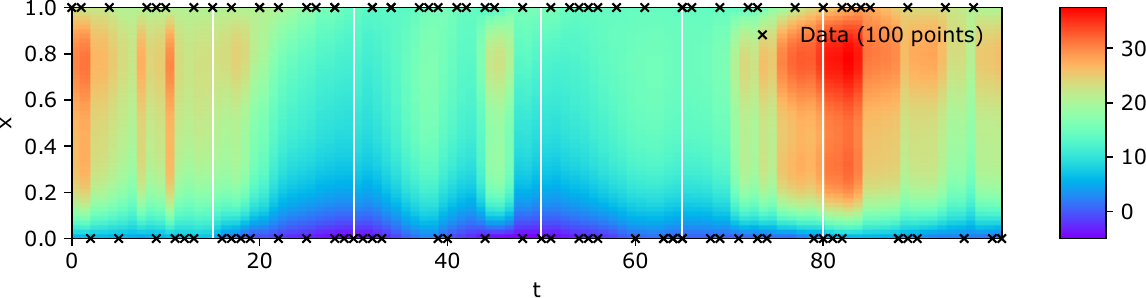}
         \caption{PINN with full pre-training.}
         \label{fig:pinn_fullpretrain_32}
    \end{subfigure}
\caption{Temperature solution for 100 hours of the heat diffusion equation using vanilla PINN, PINN with the boundary pre-training, and PINN full pre-training using 32 nodes.}
\label{fig:results_all_32}
\end{figure}

Fig. \ref{fig:results_all_32} shows the temperature solution for 100 hours of heat diffusion using vanilla PINN, PINN with boundary pre-training, and PINN with full pre-training with 32 nodes. Comparing the results to the reference solution given by Comsol, shown in Fig. \ref{fig:comsol_init}, we can see that all 3 models accurately predict the temperature solution and its variations. 

\begin{figure}[H]
\centering
     \begin{subfigure}[b]{0.9\textwidth}
    \centering
    \includegraphics[width=\textwidth]{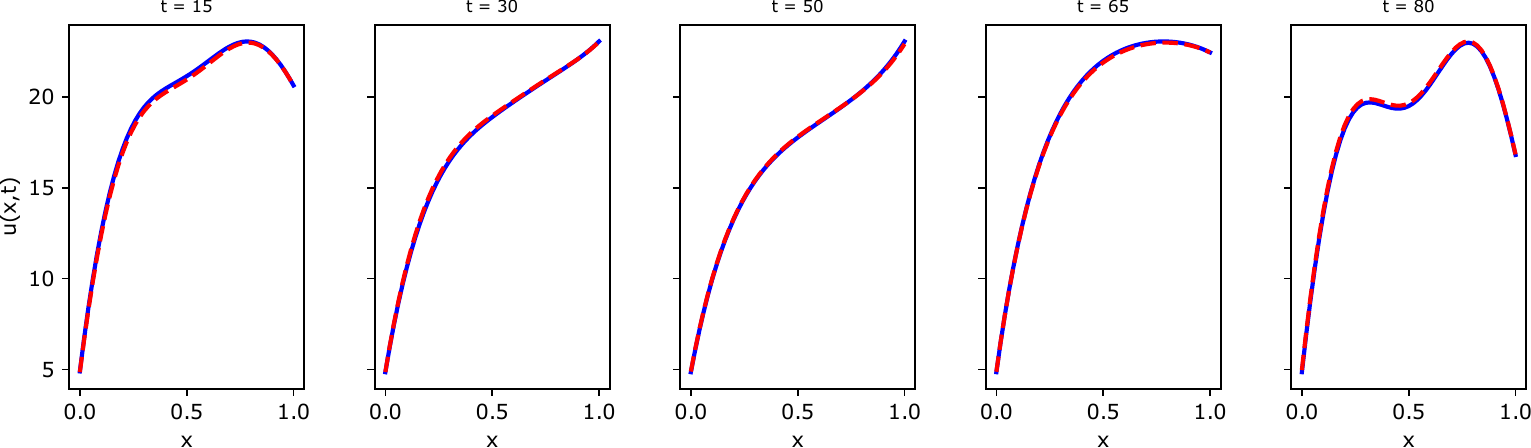}
         \caption{Vanilla PINN.}
         \label{fig:vanilla_compare_32}
    \end{subfigure}
        \hfill
     \begin{subfigure}[b]{0.9\textwidth}
    \centering
    \includegraphics[width=\textwidth]{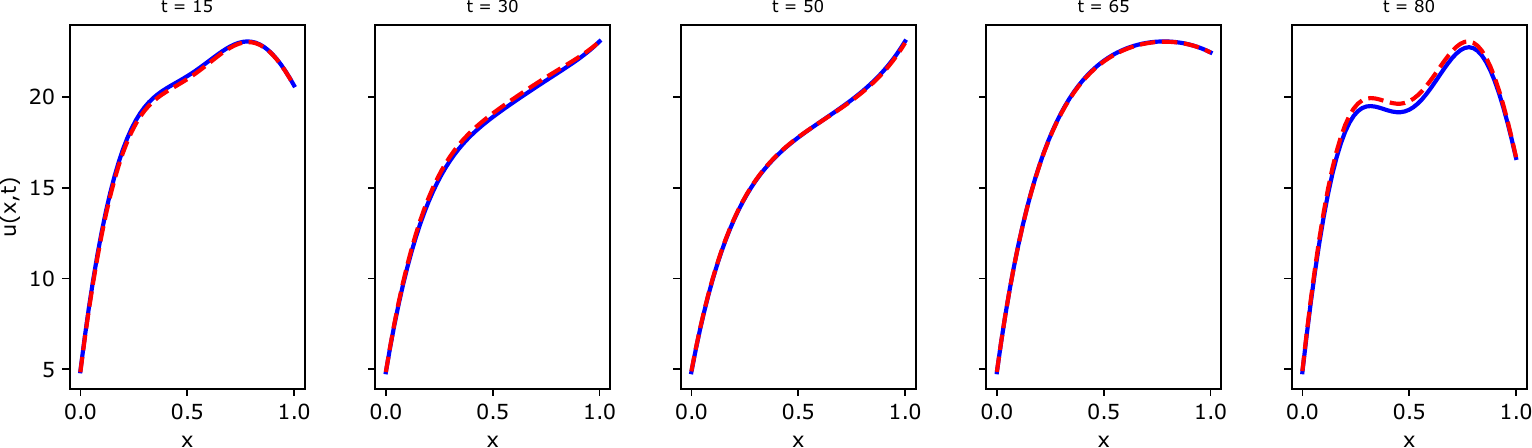}
         \caption{PINN with boundary pre-training.}
         \label{fig:boundpretrain_compare_32}
    \end{subfigure}
             \hfill
     \begin{subfigure}[b]{0.9\textwidth}
    \centering
    \includegraphics[width=\textwidth]{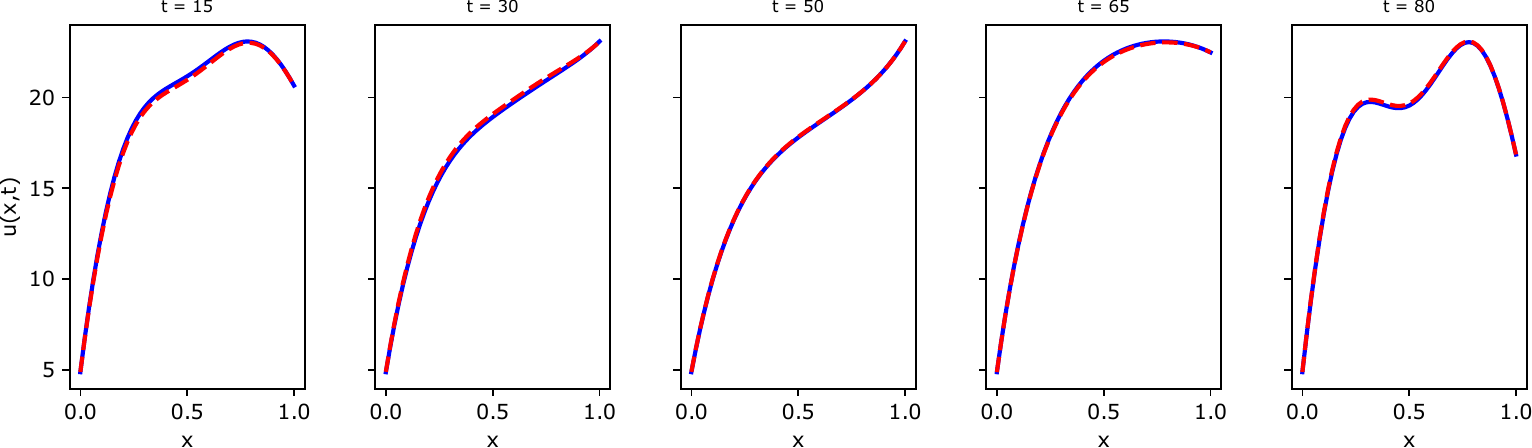}
         \caption{PINN with full pre-training.}
         \label{fig:fullpretrain_compare_32}
    \end{subfigure}
\caption{Comparison of the solution obtained using Comsol (blue lines) and vanilla PINN, PINN with the boundary pre-training, and PINN full pre-training (red-dotted lines) using 32 nodes.}
\label{fig:compare_all_32}
\end{figure}

\begin{figure}[H]
\centering
     \begin{subfigure}[b]{0.9\textwidth}
    \centering
    \includegraphics[width=\textwidth]{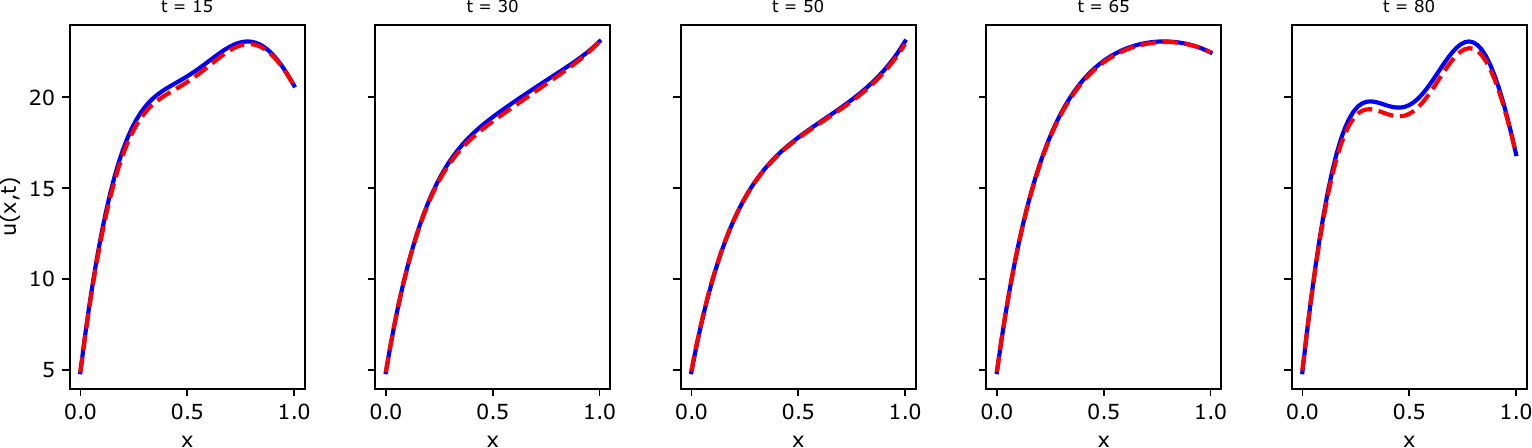}
         \caption{Vanilla PINN.}
         \label{fig:vanilla_compare_60}
    \end{subfigure}
        \hfill
     \begin{subfigure}[b]{0.9\textwidth}
    \centering
    \includegraphics[width=\textwidth]{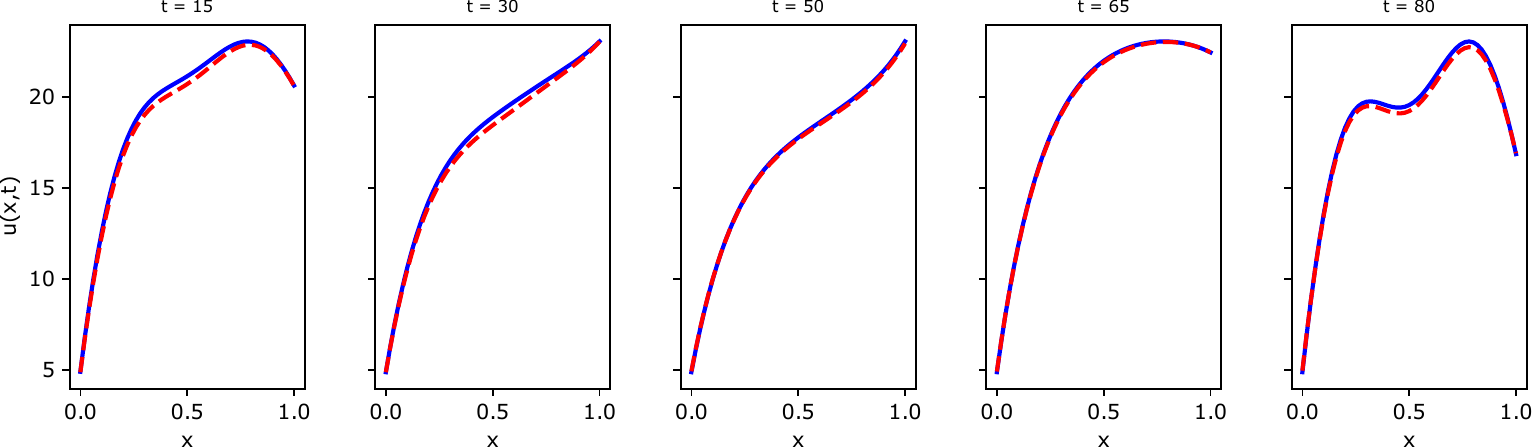}
         \caption{PINN with boundary pre-training.}
         \label{fig:boundpretrain_compare_60}
    \end{subfigure}
             \hfill
     \begin{subfigure}[b]{0.9\textwidth}
    \centering
    \includegraphics[width=\textwidth]{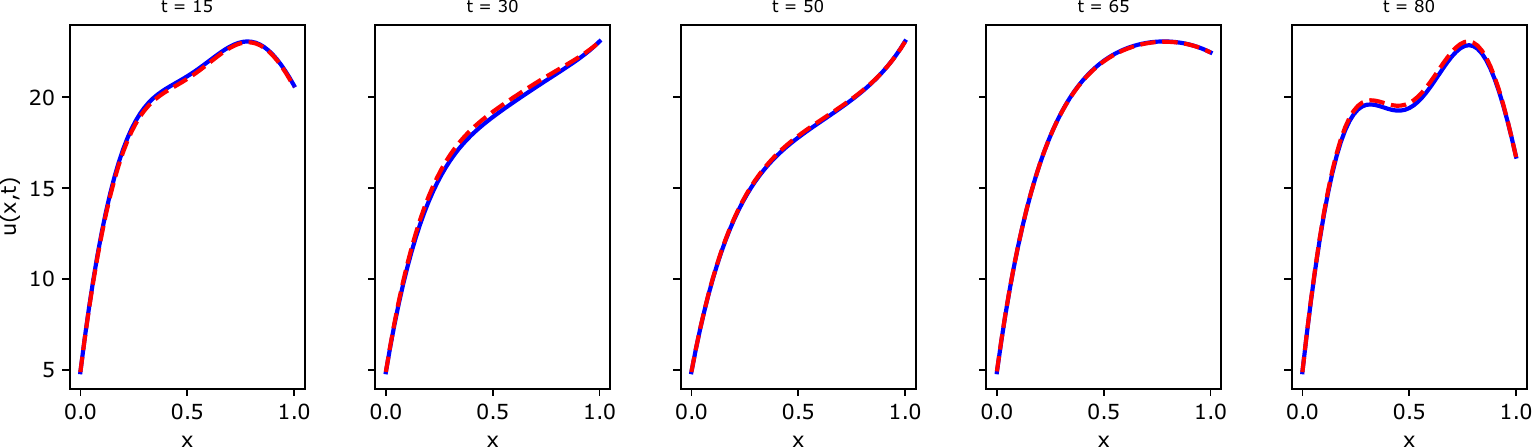}
         \caption{PINN with full pre-training.}
         \label{fig:fullpretrain_compare_60}
    \end{subfigure}
\caption{Comparison of the solution obtained using Comsol (blue lines) and vanilla PINN, PINN with the boundary pre-training, and PINN full pre-training (red-dotted lines) using 60 nodes.}
\label{fig:compare_all_60}
\end{figure}

Figs. \ref{fig:compare_all_32} and \ref{fig:compare_all_60} give a closer look at the results obtained using 32 and 60 nodes, respectively, for five instants, i.e., $t=15$, $t=30$, $t=50$, $t=65$, and $t=80$. In particular, Figs. \ref{fig:vanilla_compare_32}, \ref{fig:boundpretrain_compare_32}, and \ref{fig:fullpretrain_compare_32} show the results for vanilla PINN, PINN with boundary pre-training, and PINN with full pre-training for 32 nodes; while Figs. \ref{fig:vanilla_compare_60}, \ref{fig:boundpretrain_compare_60}, and \ref{fig:fullpretrain_compare_60} show their counterparts for 60 nodes. The blue solid lines represent the Comsol solution for all figures, while the red-dotted lines represent the PINNs solutions. All the models closely capture the temperature solutions that Comsol gave, with minor and negligible differences when using 60 nodes.

\section{Conclusion}

In this article, we investigated the effect of pre-training on the training of PINNs. To that extent, we introduced two pre-training strategies that consider solving a relaxation of the problem exactly. This relaxation is obtained by approximating the hyperbolic tangent function by the saturation. Then we investigated the impact of these pre-training algorithms on the training of the PINN. The conclusion is that using boundary pre-training - that is fitting the boundary points first - tends to significantly increase the final accuracy and decrease the training time. The second pre-training strategy which consists of also minimizing the physics-loss is actually less successful and the results are less satisfying. This is probably due to the crude approximation of the derivative of the hyperbolic tangent, which is decreasing the overall accuracy.

Finally, we can conclude that, in the heat transfer example, using a pre-training algorithm that fits the boundary points is relevant and improves the reliability of the overall training. Future works would focus on generalizing this methodology while keeping the computational burden low. Extensions to other problems will also be considered.

\section*{Acknowledgment}

This work is supported by the Vinnova Program for Advanced and Innovative Digitalisation (Ref. Num. 2023-00241) and Vinnova Program for Circular and Biobased Economy (Ref. Num. 2021-03748) and partially supported by the Wallenberg AI, Autonomous Systems and Software Program (WASP) funded by the Knut and Alice Wallenberg Foundation.